\documentclass[review]{elsarticle}

\usepackage{lineno,hyperref}
\modulolinenumbers[5]

\usepackage{amsmath}
\usepackage{changepage}
\usepackage{multirow}
\usepackage{booktabs,graphicx}
\journal{arXiv}

\begin{document}

\begin{frontmatter}

\title{A Biologically Plausible Supervised Learning Method for Spiking Neural Networks 
Using the Symmetric STDP Rule}
% \tnotetext[mytitlenote]{Fully documented templates are available in the elsarticle package on \href{http://www.ctan.org/tex-archive/macros/latex/contrib/elsarticle}{CTAN}.}

% %% or include affiliations in footnotes:
% \author[mymainaddress,mysecondaryaddress]{Yunzhe Hao}
% % \ead[url]{www.elsevier.com}
% \author[mymainaddress]{Xuhui Huang\corref{mycorrespondingauthor}}
% \author[mymainaddress,myfifthaddress]{Meng Dong}
% \cortext[mycorrespondingauthor]{Corresponding author at: Institute of Automation, Chinese Academy of Sciences(CASIA), Beijing, China\\ E-mailaddress:
% xuhui.huang@ia.ac.cn(X. Huang)}
% % \ead{support@elsevier.com}
% \author[mymainaddress,mythirdaddress,myfourthaddress]{Bo Xu}

 \author[label1,label2]{Yunzhe Hao \corref{equ} }
 \author[label1]{Xuhui Huang \corref{equ} \corref{cor}}
 
 \author[label1]{Meng Dong}
 
 \author[label1,label2,label3]{Bo Xu \corref{cor}}
 
 %\corref{equ}
 \cortext[equ]{These two authors contributed equally to this work.}
 
 \cortext[cor]{Corresponding authors at: Research Center for Brain-inspired Intelligence, Institute of Automation, Chinese Academy of Sciences, 100190 Beijing, China. E-mail addresses: xuhui.huang@ia.ac.cn (X. Huang) or xubo@ia.ac.cn (B. Xu). }
 
 \address[label1]{Research Center for Brain-inspired Intelligence, Institute of Automation, Chinese Academy of Sciences, 100190 Beijing, China}
 \address[label2]{University of Chinese Academy of Sciences, 100049 Beijing, China}
 \address[label3]{CAS Center for Excellence in Brain Science and Intelligence Technology,
 Chinese Academy of Sciences, 100190 Beijing, China} 

\begin{abstract}
Spiking neural networks (SNNs) possess energy-efficient potential due to event-based computation. However, supervised training of SNNs remains a challenge as spike activities are non-differentiable. Previous SNNs training methods can be generally categorized into two basic classes, i.e., backpropagation-like training methods and plasticity-based learning methods. The former methods are dependent on energy-inefficient real-valued computation and non-local transmission, as also required in artificial neural networks (ANNs), whereas the latter are either considered to be biologically implausible or exhibit poor performance. Hence, biologically plausible (bio-plausible) high-performance supervised learning (SL) methods for SNNs remain deficient. In this paper, we proposed a novel bio-plausible SNN model for SL based on the symmetric spike-timing dependent plasticity (sym-STDP) rule found in neuroscience. By combining the sym-STDP rule with bio-plausible synaptic scaling and intrinsic plasticity of the dynamic threshold, our SNN model implemented SL well and achieved good performance in the benchmark recognition task (MNIST dataset). To reveal the underlying mechanism of our SL model, we visualized both layer-based activities and synaptic weights using the t-distributed stochastic neighbor embedding (t-SNE) method after training and found that they were well clustered, thereby demonstrating excellent classification ability. Furthermore, to verify the robustness of our model, we trained it on another more realistic dataset (Fashion-MNIST), which also showed good performance. As the learning rules were bio-plausible and based purely on local spike events, our model could be easily applied to neuromorphic hardware for online training and may be helpful for understanding SL information processing at the synaptic level in biological neural systems.
\end{abstract}


\end{frontmatter}

% \linenumbers

\section{Introduction} \label{Sect:Intro}

Due to the emergence of deep learning technology and rapid growth of high-performance computing, artificial neural networks (ANNs) have achieved various breakthroughs in machine learning tasks~\citep{lecun2015deep,schmidhuber2015deep}. However, ANN training can be energy inefficient as communication among neurons is generally based on real-valued activities and many real-valued weights and error signals need to be transmitted in the basic error backpropagation (BP) training algorithm~\citep{rumelhart1988learning}. This can consume considerable energy for computations on central processing units (CPUs) and information transportation between CPUs and random-access memory (RAM) in traditional high-performance computers. As such, ANNs are potentially highly energy consumptive. However, information processing in the human brain is very different to that in ANNs. Neurons in biological systems communicate with each other via spikes or pulses (i.e.,~event-based communication), which allow for the asynchronous update of states only when a spike is incoming. This has the potential advantage of energy efficiency compared to the synchronous update of real-valued states at each step in ANNs. In addition, learning rules for modifying synaptic weights of realistic neurons can also be based on local spike events, e.g.,~spike-timing dependent plasticity (STDP)~\citep{bi1998synaptic,Caporale2008b,song2000competitive}, where the weight of a synapse is changed based on the spike activities of its pre-and post-synaptic neurons. This local event-based learning requires no additional energy for non-local transmission, which is required during the ANN training process. Therefore, spiking neural networks (SNNs), which can closely mimic the local event-based computations and learning of biological neurons, may be more energy efficient than ANNs, especially if SNNs are implemented onto neuromorphic platforms. Moreover, SNNs exhibit the natural ability of spatiotemporal coding of an input, and thus hold the potential advantage of efficient coding though sparse activities, particularly for continuous spatiotemporal inputs (e.g.,~optic flow), whereas ANNs require specially designed architectures to deal with temporal input (e.g.,~long short-term memory, LSTM)~\citep{hochreiter1997long,greff2016lstm} as their individual neurons generally have no intrinsic temporal characteristics and only generate output responses based on current inputs at each step. Thus, for these reasons, SNNs are considered as third generation neural network models and have attracted growing interest for exploring their functions in real-world tasks~\citep{maass2004computational,tavanaei2018deep}.

Unlike that of ANNs, SNN training is highly challenging due to the non-differentiable properties of the spike-type activity. Hence, the development of an efficient training algorithm for SNNs is of considerable importance. Much effort has been expended in the past two decades on this issue~\citep{tavanaei2018deep}, with the subsequently developed approaches generally characterized as indirect supervised learning (SL), direct SL, or plasticity-based training~\citep{tavanaei2018deep,wu2017spatio}. For the indirect SL method, ANNs are first trained and then mapped to equivalent SNNs by different conversion algorithms that transform real-valued computing into spike-based computing~\citep{cao2015spiking,hunsberger2015spiking,diehl2015fast,esser2015backpropagation,diehl2016conversion,neil2016effective,esser2016convolutional,hu2018spiking}; however, this method does not incorporate SNN learning and therefore provides no heuristic information on how to train a SNN. The direct SL method is based on the BP algorithm~\citep{wu2017spatio,lee2016training,Samadi2017Deep,tavanaei2017bp,zhangaaai2018}, e.g.,~using membrane potentials as continuous variables for calculating errors in BP~\citep{lee2016training,zhangaaai2018} or using continuous activity function to approximate neuronal spike activity and obtain differentiable activity for the BP algorithm~\citep{wu2017spatio,tavanaei2017bp}. However, such research must still perform numerous real-valued computations and non-local communications during the training process; thus, BP-based methods are as potentially energy inefficient as ANNs and also lack bio-plausibility. For plasticity-based training, synaptic plasticity rules (e.g.,~STDP) are used to extract features for pattern recognition in an unsupervised learning (USL) way~\citep{diehl2015unsupervised}. Due to the nature of spontaneous unsupervised clustering of synaptic plasticity, this method requires an additional supervised module for recognition tasks. Three supervised modules have been used in previous studies: (1) a classifier (e.g.,~support vector machine, SVM)~\citep{kheradpisheh2017stdp}, (2) a label statistical method outside the network~\citep{diehl2015unsupervised}, and (3) an additional supervised layer~\citep{Beyeler2013Categorization,hu2017stdp,shrestha2017stable,mozafari2018combining}. In our opinion, a neural network model with biological plausibility must meet the following basic characteristics. Firstly, the single neuron model must integrate temporal inputs and generate pulses or spikes as outputs. Secondly, the computation processes of training and inference must be completely spike-based. Finally, all learning rules must be based on experiments, and should not be violated (obviously contrary to) by experiments or artificially designed. The first two supervised modules are bio-implausible due to the need of computation outside the SNNs~\citep{kheradpisheh2017stdp,diehl2015unsupervised}. The last supervised module has the potential of bio-plausibility, but existing supervised SNN models have either adopted artificially modified STDP rules~\citep{Beyeler2013Categorization,hu2017stdp,shrestha2017stable,mozafari2018combining} or exhibited poor performance~\citep{Beyeler2013Categorization}. Currently, therefore, truly bio-plausible SNN models that can accomplish SL and achieve high-performance pattern recognition are lacking. Moreover, although great progress has been made in understanding the physiological mechanisms for the modification of synapses at the microscopic level~\citep{bi1998synaptic,Caporale2008b}, how teacher learning at the macroscopic behavioral level is realized by changes in synapses at the microscopic level, i.e.,~the mechanism of SL processing in the brain, is still far from clear.

\begin{figure}[htb]
    \centering
    %\fbox{\rule[-.5cm]{0cm}{4cm} \rule[-.5cm]{4cm}{0cm}}
    \includegraphics[scale=1]{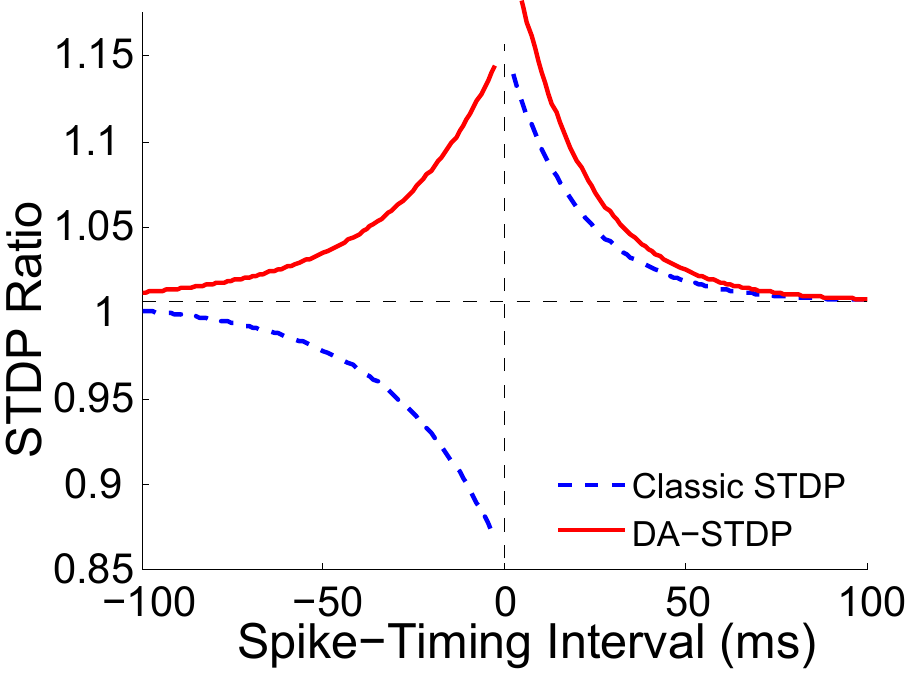}
    \caption{Schematic of the classic STDP \cite{bi1998synaptic} and DA-STDP \cite{Zhang2009Gain,brzosko2015retroactive}.}
     \label{DA-STDP_and_STDP}
\end{figure}

In this study, we proposed a novel bio-plausible SL method for training SNNs based on biological plasticity rules. We introduced the dopamine-modulated STDP (DA-STDP) rule, a new type of symmetric STDP (sym-STDP), for pattern recognition. The DA-STDP rule has been observed in several different experiments in the hippocampus and prefrontal cortex \cite{Zhang2009Gain,Ruan2014,brzosko2015retroactive}, where the modification of synaptic weight is always incremental if the interval between the pre- and post-synaptic spike activities is within a narrow time-window when dopamine (DA) is present. The differences between the sym- or DA-STDP and classic STDP rules are shown in Figure~\ref{DA-STDP_and_STDP} \cite{bi1998synaptic,Zhang2009Gain,brzosko2015retroactive}. While the sym-STDP rule has been used previously \cite{masuda2007formation,tanaka2009cmos,serrano2013stdp,mishra2016symmetric}, this is the first time it has been applied for SL. In our proposed model, a three-layer feedforward SNN was trained by DA-STDP combined with synaptic scaling \cite{turrigiano2008the,turrigiano1998activity,effenberger2015self-organization} and dynamic threshold, which are two homeostatic plasticity mechanisms for stabilizing and specializing the network response under supervised signals. Two different training methods were used in our SNN model, i.e., training two-layer input synaptic weights simultaneously and training the SNN layer-by-layer. Our model was tested in the benchmark handwritten digit recognition task (MNIST dataset) and achieved high performance under the two training methods. We also evaluated the model using the MNIST-like fashion product dataset (Fashion-MNIST), which also showed good classification performance. These results thus highlighted the robustness of our model.

\begin{figure}[htb]
        \centering
        %\fbox{\rule[-.5cm]{0cm}{4cm} \rule[-.5cm]{4cm}{0cm}}
        \includegraphics[scale=0.8]{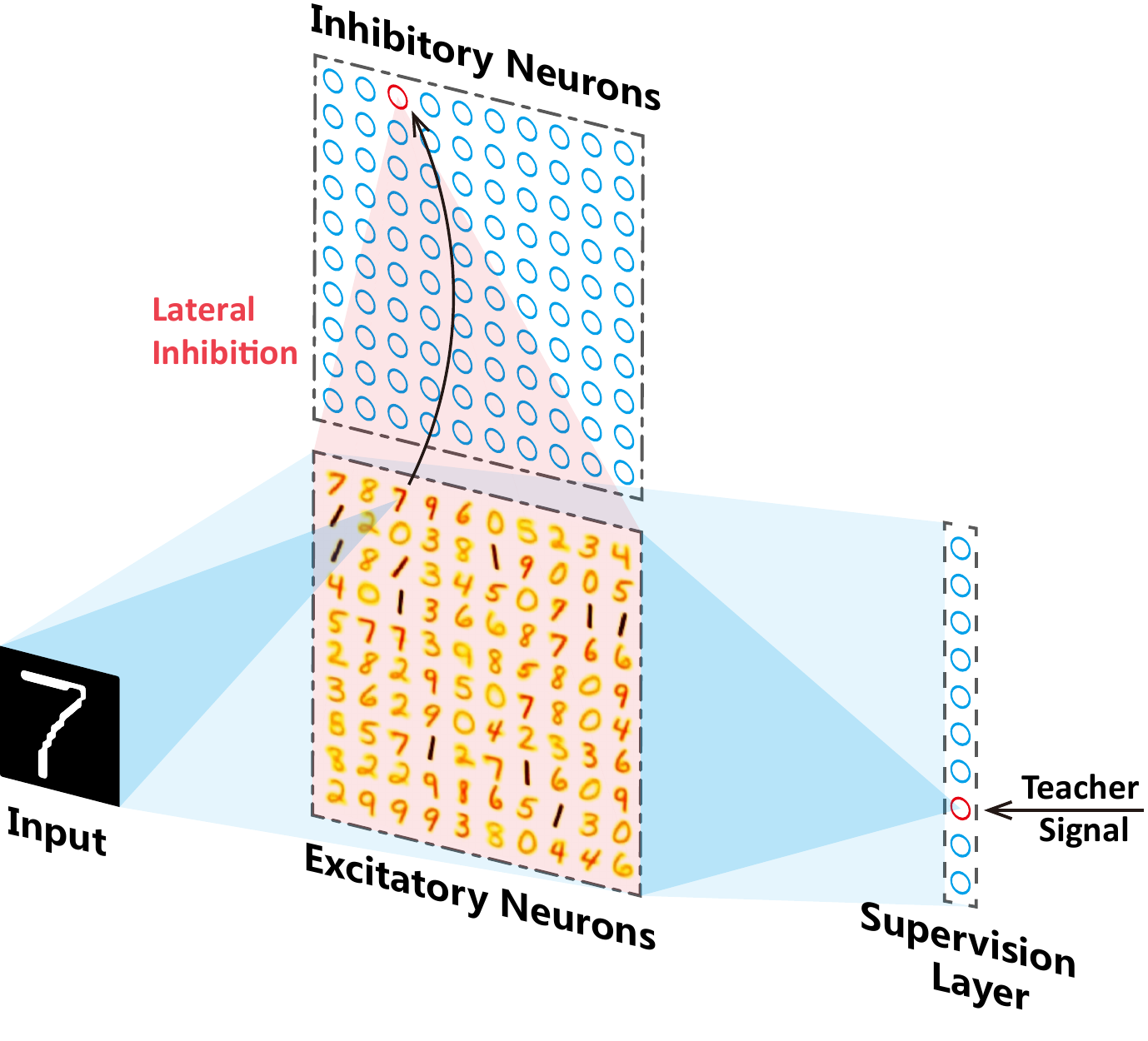}
        \caption{Network structure.}
         \label{network_structure}
\end{figure}

% \begin{itemize}
% \item document style
% \item baselineskip
% \item front matter
% \item keywords and MSC codes
% \item theorems, definitions and proofs
% \item lables of enumerations
% \item citation style and labeling.
% \end{itemize}

\section{Network architecture and neuronal dynamics}
We constructed a three-layer feedforward spiking neural network for SL, which included an input layer, hidden layer, and SL layer (Figure~\ref{network_structure}). The structure of the first two layers was inspired by the USL model of Diehl and Cook \cite{diehl2015unsupervised}. Input patterns were coded as Poisson spike processes with firing rates proportional to the intensities of the corresponding pixels. The Poisson spike trains were then fed to the excitatory neurons in the hidden layer with all-to-all connections. The dark blue shaded area in Figure~\ref{network_structure} shows the input connection to a specific neuron. The connection from the excitatory to inhibitory neurons was one-to-one. An inhibitory neuron only received input from the corresponding excitatory neuron at the same position in the map and inhibited the remaining excitatory neurons. All excitatory neurons were fully connected to the SL layer. In the SL layer, neurons fired with two different modes during the training and testing processes. During the SL training period, the label information of the current input pattern was converted to a teacher signal in a one-hot coding scheme by the 10 SL neurons. Only one SL neuron was pushed to fire as a Poisson spike process, with the remaining SL neurons maintained in the resting state. In the testing mode, all SL neurons fired according to inputs from the hidden layer.

%Since the category labels are presented to the supervision layer, our learning is supervised. 
%There is no algorithms to propagate supervision information back like backpropagation, so the supervised is weak.

We used the leaky integrate-and-fire (LIF) model and set the parameters within bio-plausible ranges. The resting membrane potential $E_{\text{rest}}$ was set at $-65$ mV and the equilibrium potentials of the excitatory synapses $E_{\text{E}}$ and inhibitory synapses $E_{\text{I}}$ were set to $0$ mV and $-100$ mV, respectively. The time constant of membrane potential damping $\tau$ was equal to $100~{\rm ms}$. The membrane potential $V$ dynamics of the neurons can be described by~\cite{diehl2015unsupervised} and \cite{vogels2011inhibitory}:

\begin{eqnarray}
        \label{eq:LIF}
        \tau \frac{dV}{dt} = (E_\text{rest}-V)+g_\text{E} (E_\text{E}-V)+g_\text{I} (E_\text{I}-V) 
\end{eqnarray}
where $g_\text{E}$ and $g_\text{I}$ are the total excitatory and total inhibitory conductances, respectively.

Both $g_\text{E}$ and $g_\text{I}$ have a similar dynamic equation. They are dependent on the number of excitatory and inhibitory synapses ($N_\text{E}$ and $N_\text{I}$) and input synapses (in-synapses) weights ($w_i^\text{E}$ and $w_i^\text{I}$) of excitatory and inhibitory synapses, respectively. Both of the time constants of synapse conductance damping $\tau_{g_\text{E}}$ and $\tau_{g_\text{I}}$ were equal to $1$ ms. Thus, $g_\text{E}$ and $g_\text{I}$ in Eq.~\ref{eq:LIF} can be described by the following equations \cite{diehl2015unsupervised,vogels2011inhibitory}:
\begin{eqnarray}
        \label{eq:g_E}
        \tau_{g_\text{E}} \frac{dg_\text{E}}{dt} = -{g_\text{E}}+ \sum_{i = 1}^{N_\text{E}} {\sum_{k}{w_i^\text{E}\delta(t-t_i^k)}} \\
        \label{eq:g_I}
        \tau_{g_\text{I}} \frac{dg_\text{I}}{dt} = -{g_\text{I}}+ \sum_{i = 1}^{N_\text{I}} {\sum_{k}{w_i^\text{I}\delta(t-t_i^k)}} 
\end{eqnarray}
where $t_i^k$ is the $k$th spike time from the $i$th neuron.

In our neuron model, once the membrane potential exceeds the threshold voltage, the membrane potential is returned to the reset voltage $E_\text{reset}$ ($-65$ mV), and the neuron will not fire again in the refractory time $T_\text{refractory}$ ($2$ ms). The firing threshold is not static, with a dynamic threshold (i.e., homeostatic plasticity) mechanism instead adopted. The dynamic threshold is the intrinsic plasticity of a neuron, and is found in different neural systems \cite{yeung2004synaptic,sun2009experience,zhang2003other,pozo2010unraveling,cooper2012bcm}. Here, it was introduced to generate a specific response to a class of input patterns for each excitatory neuron in the hidden layer \cite{zhang2003other,pozo2010unraveling,diehl2015unsupervised}, otherwise single neurons can dominate the response pattern due to their enlarged input synaptic (in-synaptic) weights and lateral inhibition. The membrane potential threshold $V^{\text{th}}$ was composed of two elements (Eq.~\ref{eq:threshold_add}), i.e., the constant ($V^{\text{th}}_{\text{const}} = -72$ mV) and dynamic variable ($\theta$) parts. $\theta$ increases slightly when the neuron fires, otherwise it decays exponentially. As a neuron will not (or barely) fire when $\theta$ is too large, which can negatively impact model performance, we adopted a dynamical increment to slow $\theta$ growth gradually. Therefore, $V^{\text{th}}$ can be described as:

\begin{eqnarray}
        \label{eq:threshold_add}     
           V^{\text{th}} &=& V^{\text{th}}_{\text{const}} + \theta           \\
        \label{eq:threshold_intrinsic_plasticity}    
        \tau_{\theta} \frac{d\theta}{dt} &=& - \theta + \frac{\theta_{\text{initial}}} { \vert 2\theta - \theta_{\text{initial}} \vert }\sum_{k} {\alpha \delta(t-t_k) }
\end{eqnarray}
where $\tau_{\theta}$ is the time constant of $\theta$, with the initial value of $\theta$ ($\theta_{\text{initial}}$) set to $20$ mV, $\alpha$ is the maximum value of the increment and $\frac{\theta_{\text{initial}}} { \vert 2\theta - \theta_{\text{initial}} \vert }$ is its dynamical scaling factor, and $t_k$ is the $k$-th firing time.

Synaptic weights were modified according to the two biological plasticity rules DA-STDP and synaptic scaling. Dopamine is an important neuromodulator and plays a critical role in learning and memory processes \cite{tritsch2012dopaminergic}. Here, inspired by the DA-STDP found in different brain areas, such as the hippocampus and prefrontal cortex \cite{Zhang2009Gain,Ruan2014,brzosko2015retroactive}
% \cite{Zhang2009Gain,Chiu2010,Edelmann2011,Ruan2014}
(Figure~\ref{DA-STDP_and_STDP}), we hypothesized that DA can modulate changes in synaptic weights according to the DA-STDP rule during the SL process. The weights increment $\Delta W$ under the phenomenological model of DA-STDP can be expressed as \cite{Zhang2009Gain,brzosko2015retroactive}:
\begin{eqnarray}
        \Delta W =
        \begin{cases}
                %A_+ \sum_{k} \exp(\frac{- \Delta t}{\tau_+}), & \Delta t > 0 \\
                %A_- \sum_{k} \exp(\frac{\Delta t}{\tau_-}), & \Delta t < 0    
                A_{+} \exp(\frac{-\Delta t}{\tau_{+}}), & \Delta t > 0 \\
                A_{-} \exp(\frac{\Delta t}{\tau_{-}}), & \Delta t < 0    
        \end{cases}
        \label{eq:DA-STDP}
\end{eqnarray}
where $\Delta t$ is the time difference between the pre- and post-synaptic spikes, $\tau_{+}$ and $\tau_{-}$ are the time constants of positive and negative phases for $\Delta t$, respectively, and $A_{+}$ and $A_{-}$ are learning rates.
%\begin{eqnarray}
%       W(x) =
%       \begin{cases} 
%               A_+ exp(-\Delta t /\tau_+),  & \Delta t \geq 0 \\
%               -A_- exp(-\Delta t /\tau_-), & \Delta t < 0    \\
%       \end{cases}                     
%       \label{eq:Classic STDP}
%\end{eqnarray}
%\begin{eqnarray}
%       W_2(\Delta t) =
%       \begin{cases} 
%               A_2^+ exp(-\Delta t /\tau_+), & \Delta t \geq 0 \\
%               -A_2^- exp(\Delta t /\tau_-), & \Delta t < 0    \\
%       \end{cases}
%       \label{eq:Pair-based STDP}
%\end{eqnarray}
%
%\begin{eqnarray}
%       W_3(\Delta t_1,\Delta t_2) = A_3^+ exp(-\Delta t_1 /\tau_+) exp(-\Delta t_2 /\tau_y), & \Delta t_1 \geq 0,\Delta t_2 \geq 0 \\
%       \Delta t = t^{post} - t^{pre} \quad \Delta t_1 = t^{post} - t^{pre} \ \mbox{and} \ \Delta t_2 = t^{post} - t^{\prime pre}
%       \label{eq:Triplet STDP}
%\end{eqnarray}

As DA-STDP can only increase synapse strength, the synaptic scaling plasticity rule was introduced to generate a competition mechanism among all in-synapses of a neuron in the hidden layer (only for excitatory neurons) and SL layer. Synaptic scaling is a homeostatic plasticity mechanism observed in many experiments~\citep{pozo2010unraveling,turrigiano2008the,davis2006homeostatic}, especially in visual systems~\citep{maffei2008multiple,keck2013synaptic,hengen2013firing} and the neocortex~\citep{turrigiano1998activity}. Here, synaptic scaling was conducted after the pattern was trained and synapse strength was normalized according to the following equation~\citep{diehl2015unsupervised}:
\begin{eqnarray}
     w^{'} = w \frac{\beta  N_\text{in}}{\sum{w}} 
        \label{eq:synaptic_scaling}
\end{eqnarray}
where $N_{\text{in}}$ is the number of all in-synapses of a single neuron and $ \beta $ with $\beta \in (0,1) $ is the scaling factor.
\section{Recognition performance for the MNIST task}
%\subsection{Performance under different training methods }
%\subsection{Convergence property }
Our SNN model was trained on the MNIST dataset (training set: 60\,000 samples; test dataset: 10\,000 samples) (http://yann.lecun.com/exdb/mnist/) using two training methods, i.e.,~simultaneous and layer-by-layer training. For simultaneous training, the in-synapses of the hidden and SL layers were updated simultaneously during the training process, whereas for layer-by-layer training, the hidden layer was trained first, then the SL layer was trained with all in-synaptic weights of the hidden layer fixed. In the inference stage, the most active SL layer neuron was regarded as the inference label of the input sample. Error rate on the test dataset was used to evaluate performance. No data preprocessing was conducted, and we used a SNN simulator (GeNN)~\citep{yavuz2016genn} to simulate all experiments.

For the model parameters, we set the simulation time step to $0.5~{\rm ms}$ and the presentation time of a single input sample to $350~{\rm ms}$, followed by a resting period of $150~{\rm ms}$. Both $\tau_{\theta}$ and $\alpha$ were tuned according to network size. The in-synaptic and output synaptic (out-synaptic) weights of the excitatory neurons in the hidden layer were in the ranges of $[0, 1]$ and $[0, 8]$, respectively. All initial weights were set to corresponding maximum weights multiplied by uniform distributed values in the range $[0, 0.3]$. The factor for synaptic scaling $\beta$ was set to 0.1. The firing rates of the input neurons were proportional to the intensity of the pixels of the MNIST images~\citep{diehl2015unsupervised}. We set maximum rates to 63.75~Hz after dividing the maximum pixel intensity of 255 by 4. When less than five spikes were found in the excitatory neurons of the hidden layer during 350~ms, the maximum input firing rates were increased by 32~Hz. The firing rates of the SL layer neurons were 200~Hz or 0~Hz according to the one-hot code in the SL training period.

To demonstrate the power of the proposed SL method for different network sizes, we compared the results of our SL algorithm to the `Label Statistics' algorithm used in previous study~\citep{diehl2015unsupervised}. In this algorithm, an additional module first calculates the most likely representation of a neuron for a special class of input patterns, labels the neuron to the class during the training process, and uses the maximum mean firing rates among all classes of labeled neurons for inference in the test process.

\begin{figure*}[htb]
        \centering
        %\fbox{\rule[-.5cm]{0cm}{4cm} \rule[-.5cm]{4cm}{0cm}}
        \includegraphics[scale=0.8]{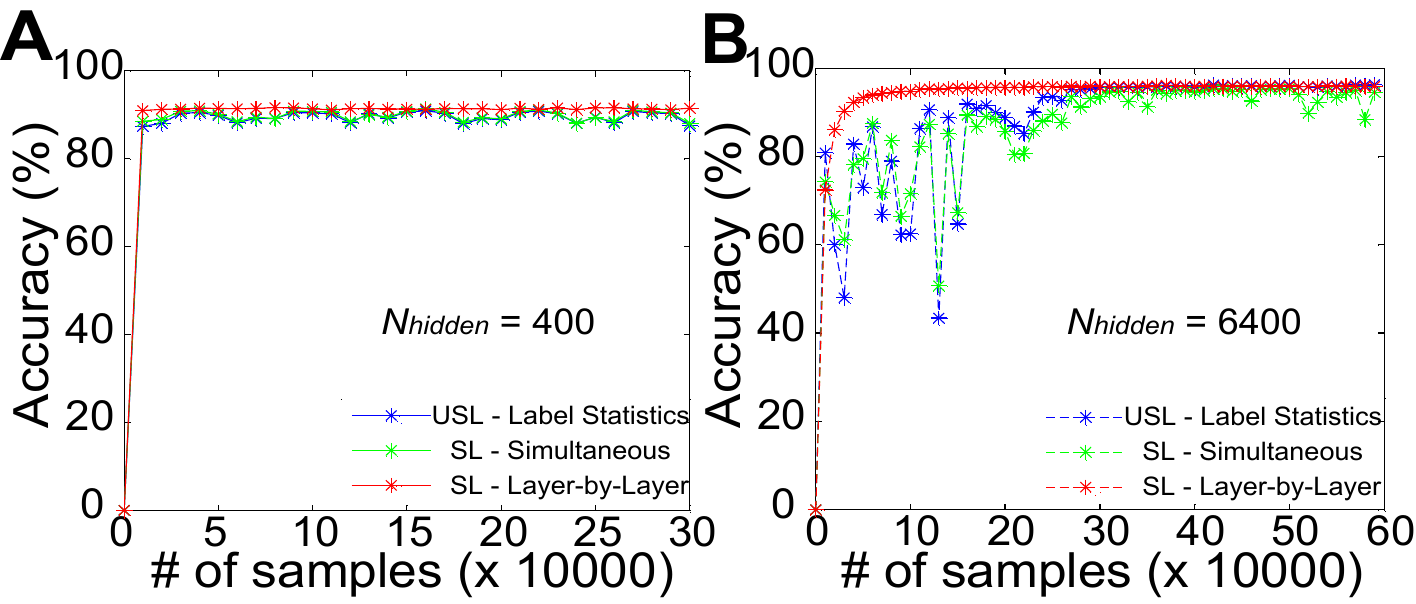}
        \caption{Convergence property of networks with sizes $N_\text{hidden}$ = 400 (A) and 6400 (B). `Layer-by-Layer' curve in each figure represents SL performance under layer-by-layer training using STDP to train in-synaptic weights of the hidden layer in an unsupervised manner and then train out-synaptic weights in a supervised manner.}
        \label{converged_perf}
\end{figure*}

\begin{figure}[htb]
        \centering
        %\fbox{\rule[-.5cm]{0cm}{4cm} \rule[-.5cm]{4cm}{0cm}}
        \includegraphics[scale=0.9]{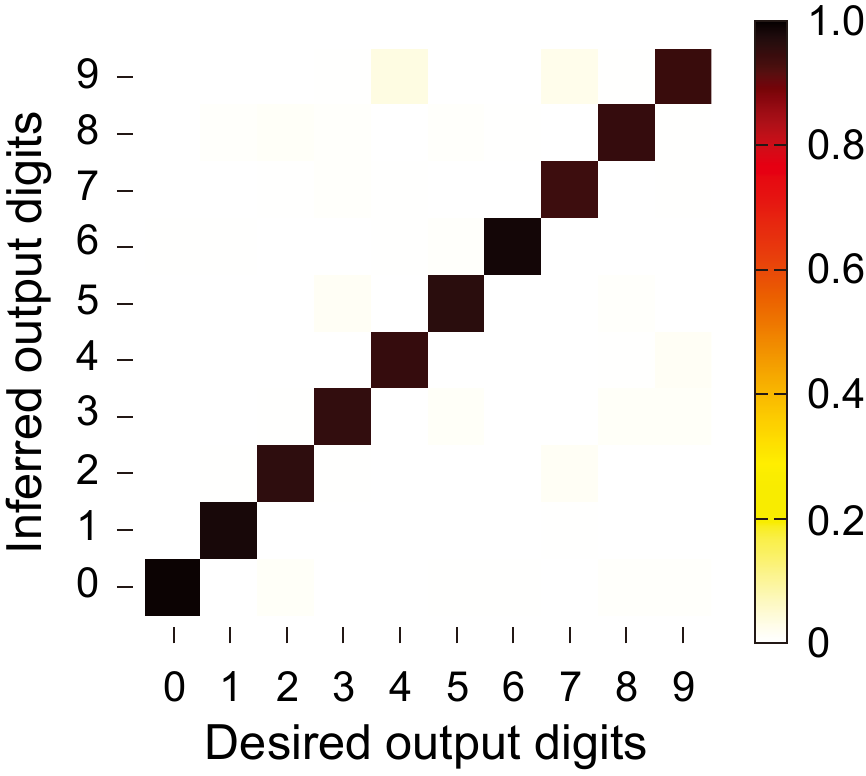}
        \caption{Confusion matrix of test dataset results in the SL layer for the network with size $N_\text{hidden}$ = 6400. Darker pixel indicates stronger consistency between desired (label) and inferred (real) outputs. Data were obtained for re-trained SL, as shown in Figure~\ref{converged_perf}B.}
        \label{confusion_m_SL_6400}
\end{figure}

We trained the model with different epochs of the training dataset for different network sizes (3, 5, 7, 10, and 20 epochs for the excitatory neuron number of the hidden layer $N_\text{hidden} = $ 100, 400, 1600, 6400, and 10000, respectively). During the training process, the network performances for each training set of 10 000 samples were estimated by the test dataset. Taking network sizes $N_\text{hidden} = $ 400 and 6400 as examples, classifying accuracies converged quickly under the two training methods (Figure~\ref{converged_perf}). Figure~\ref{confusion_m_SL_6400} shows very high consistency between the desired (label) and inferred (real) outputs in the SL layer.

The results for the different learning methods are summarized in Table~\ref{tab:perf-for-diff-nw}. Our two SL methods outperformed the `Label Statistics' method for all small-scale networks ($N_\text{hidden} = 100, 400, \text{and} \ 1600$). In addition, the `Layer-by-Layer' training method outperformed `Label Statistics' for all network scales. The best performance of our SL model (96.73$\%$) was achieved in the largest network under the `Layer-by-Layer' training method. These results indicate that a SNN equipped with biologically realistic plasticity rules can achieve good SL by pure spike-based computation. 

\begin{table*}[htb]
        \begin{adjustwidth}{}{}
        \scalebox{0.66}{
        \begin{tabular}{c c c c c c}
          \toprule
          Network Size ($N_\text{hidden}$)&100&400&1600&6400&10000   \\
          \midrule
          Simultaneous SL & $83.67\%\pm0.30\%$ & $91.13\%\pm0.05\%$ & $91.22\%\pm0.13\%$ & $95.89\%\pm0.14\%$ & $96.35\%\pm0.09\%$ \\
          \midrule
          Layer-by-layer SL &  $83.57\%\pm0.33\%$ &$91.41\%\pm0.15\%$ & $91.82\%\pm0.41\%$ & $96.28 \%\pm 0.10\%$ & $96.73 \%\pm 0.11\%$ \\
          \midrule
          Label Stat. & $83.00\%\pm0.50\%$ & $91.02\%\pm0.23\%$ & $91.19\%\pm0.32\%$ & $96.02 \%\pm 0.08 \%$ & $96.52 \%\pm 0.13 \%$ \\
          \bottomrule
        \end{tabular}
        }
        \caption{Performance for different sized networks with different classification and training methods. `Simultaneous' and `Layer-by-layer' are two training methods for SL. `Label Stat.' represents classification by the `Label Statistics' method in the hidden layer, for details please refer to \cite{diehl2015unsupervised}. We conducted five trials for each case and reported average accuracy in each case. For network size $N_\text{hidden} = 100, 400, 1600, 6400$, and $10000$, in Equation~\ref{eq:threshold_intrinsic_plasticity}, $\tau_{\theta} = 6\times 10^{6}, 6\times 10^{6}, 8\times 10^{6}, 2\times 10^{7}$ and $2\times 10^{7}$, $\alpha = 8.4\times 10^{5}, 8.4\times 10^{5}, 1.12\times 10^{6}, 2\times 10^{6}$ and $2\times 10^{6}$, respectively.}
        \label{tab:perf-for-diff-nw}
        \end{adjustwidth}
\end{table*}

\section{Visualization of model clustering ability}
\begin{figure*}[htb]
        \centering
        %\fbox{\rule[-.5cm]{0cm}{4cm} \rule[-.5cm]{4cm}{0cm}}
        \includegraphics[scale=0.65]{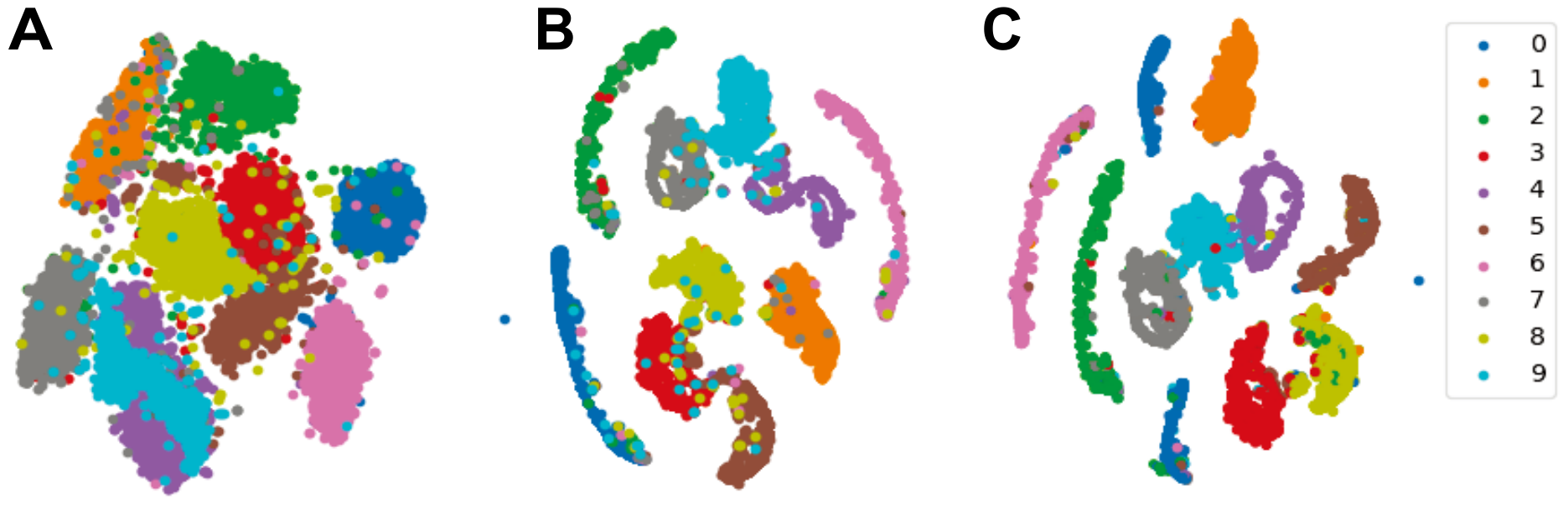}
        \caption{Visualization of original digit patterns (A), spike activities of hidden layer (B), and spike activities of SL layer (C) for the MNIST test dataset using the t-SNE method. Each dot represents a digit sample and is colored by its corresponding label information. Activities were obtained from the case of the SL-Layer-by-Layer for $N_\text{hidden}$ = 6400 (shown in Figure~\ref{converged_perf}).}
        \label{fig:Out-activities_by_t_SNE}
\end{figure*}

\begin{figure}[htb]
        \centering
        %\fbox{\rule[-.5cm]{0cm}{4cm} \rule[-.5cm]{4cm}{0cm}}
        \includegraphics[scale=0.55]{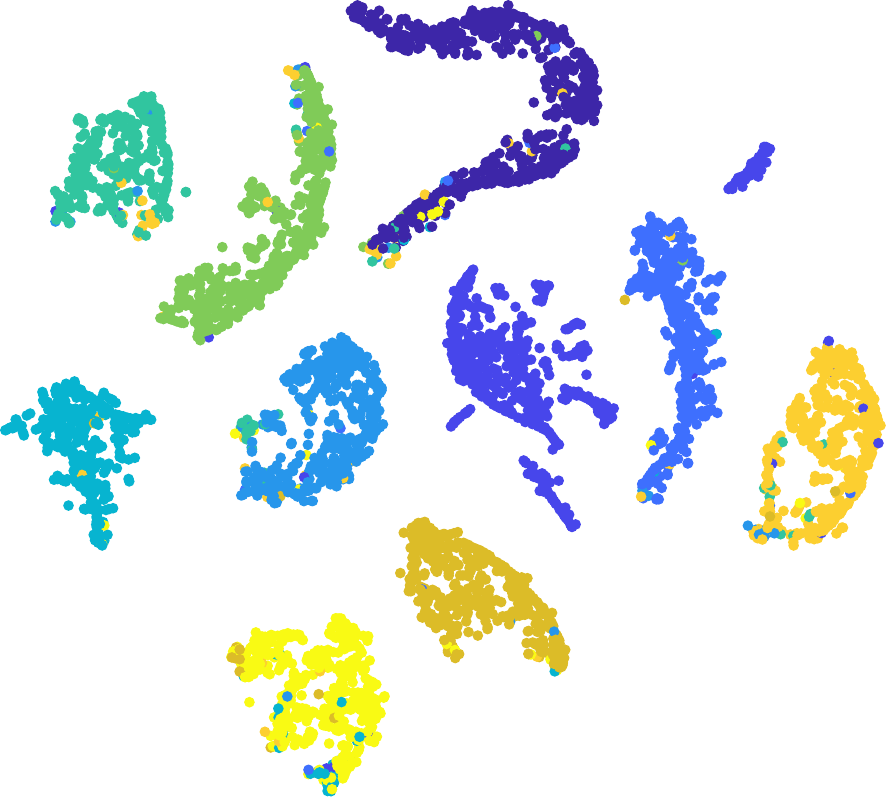}
        \caption{Visualization of the clustering ability of output synapses (out-synapses) of excitatory neurons in the hidden layer ($N_\text{E}$ = 6400) to SL layer using the t-SNE method. Each dot represents an excitatory neuron in the hidden layer and is colored by its label using `Label Statistics'. Clustering of the out-synaptic weights of the excitatory neurons is highly consistent with the clustering of their labels. Synaptic weights used here were from the case in Figure~\ref{fig:Out-activities_by_t_SNE}.}
        \label{fig:Out-synaptic_weights_by_t_SNE}
\end{figure}

To demonstrate the underlying mechanisms of our SL model in pattern recognition tasks, we adopted the t-distributed Stochastic Neighbor Embedding (t-SNE) method \cite{maaten2008visualizing} to reveal the model's clustering ability. The t-SNE is a popular nonlinear dimensionality reduction method widely used for visualizing high-dimensional data in low-dimensional space (e.g., two or three dimensions). We visualized the original digit patterns (Figure~\ref{fig:Out-activities_by_t_SNE}A), spike activities of the hidden layer (Figure~\ref{fig:Out-activities_by_t_SNE}B), and spike activities of the SL layer (Figure~\ref{fig:Out-activities_by_t_SNE}C) for all samples in the test dataset. The separability of the output information of the three layers in our model increased from the input to SL layer, indicating that the SL layer served as a good classifier after training.
To demonstrate why our SL method achieved effective clustering for the hidden layer outputs, we also applied t-SNE to reduce the dimensions of the out-synaptic weights of the excitatory neurons. As shown in Figure~\ref{fig:Out-synaptic_weights_by_t_SNE}, the clustering of the out-synaptic weights of the excitatory neurons was highly consistent with the clustering of their label information using the `Label Statistics' method. This explains why our SL method achieved comparatively good performance as the `Label Statistics' method for the classification task, although our model did not require `Label Statistics' computation outside the network to calculate the most likely representation of a hidden neuron \cite{diehl2015unsupervised}, instead realizing SL based solely on computation within the network.

\section{Comparison with other SNN models}
Current SNN models for pattern recognition can be generally categorized into three classes: that is, indirect training \cite{cao2015spiking,hunsberger2015spiking,diehl2015fast,esser2015backpropagation,diehl2016conversion,neil2016effective,esser2016convolutional,hu2018spiking}, direct SL training with BP \cite{wu2017spatio,Beyeler2013Categorization,lee2016training,Samadi2017Deep,tavanaei2017bp,zhangaaai2018,Lee2018}, and plasticity-based unsupervised training with supervised modules \cite{querlioz2013immunity,diehl2015unsupervised,kheradpisheh2017stdp}. Table~\ref{tab:SNN-models-comparison} summaries several previous SNN models trained and tested using the full training and testing sets of the MNIST dataset.
%In our model, we used bio-plausible neuroplasticity rules (i.e., sym-STDP and synaptic scaling) and local spike-based computation to accomplish SL, without the need of backpropagation or error computation. Thus, the proposed method provides a novel model framework to achieve efficient SL in SNNs.

{\bf{Comparison with SNN models trained using BP}}. 
In previous studies with indirect training, ANNs were trained using the BP algorithm based on activity rates and transformed to corresponding equivalent SNNs based on firing rates. Although their performances were very good, they ignored the temporal evolution of SNNs and spike-based learning processes. Thus, indirect training provides very little enlightenment on how SNNs learn and encode different features of inputs. For other studies using direct SL training, most adopted the BP algorithm and calculated errors based on continuous variables, e.g., membrane potentials (voltage), currents, or activity rates, to approximate spike activities and achieve SL \cite{wu2017spatio,lee2016training,Samadi2017Deep,tavanaei2017bp,zhangaaai2018,Lee2018}. For example, Zhang et al. proposed a voltage-driven plasticity-centric SNN for SL \cite{zhangaaai2018}, with four learning stages required for training, i.e., equilibrium learning and voltage-based STDP-like learning rule for USL as well as voltage-based BP for SL; however, this resulted in the model being highly dissimilar to biological neuronal systems. Lee et al. pre-trained a multi-layer SNN system by STDP in an USL way for optimal initial weights and then used current-based BP to re-train all-layer weights in a supervised way \cite{Lee2018}; however, this also resulted in the model being bio-implausible due to the use of the BP algorithm.

{\bf{Comparison with STDP-based models without BP}}.
Several studies have also proposed STDP-based training methods without BP. These previous models adopted STDP-like plasticity rules for USL and required a special supervised module for SL, e.g., a classifier (SVM) \cite{kheradpisheh2017stdp}, artificial label statistics outside the network \cite{querlioz2013immunity,diehl2015unsupervised}, or additional supervised layer \cite{Beyeler2013Categorization,hu2017stdp,shrestha2017stable,mozafari2018combining}. However, the first two supervised modules are bio-implausible because their computing modules are outside the SNNs, resulting in the SNNs having no direct relationship with SL \cite{querlioz2013immunity,diehl2015unsupervised,kheradpisheh2017stdp}. For example, Beyeler et al. adopted a calcium concentration-based STDP for SL \cite{Beyeler2013Categorization}, which showed considerably poorer performance than that of our model. Hu et al. used an artificially modified STDP with a special temporal learning phase for SL \cite{hu2017stdp}; however, their STDP rule was artificially designed and its internal mechanism was not well explained. Shrestha et al. also adopted a specially modified STDP rule with exponential weight change and extended depression window for SL in a SNN, with a similar supervised module as ours, but performance was relatively poor (less than 90$\%$) \cite{shrestha2017stable}. Mozafari et al. used a mixed classic STDP and anti-STDP rule to generate reward-modulated STDP with a remote supervised spike for SL, but they were not able to provide biological evidence to explain this type of learning \cite{mozafari2018combining}.

{\bf{Detail comparison with STDP-based SNN model by Diehl and Cook}}.
Our model was inspired by the STDP-based SNN method proposed by Diehl and Cook \cite{diehl2015unsupervised}. In their two-layer SNN, STDP is used to extract features by USL and an additional computing module outside the SNN is used for label statistics in the training process and classification in the testing process.
%The additional module first calculates the most likely representation of a neuron for a special class of input patterns, labels the neuron to the class during the training process, and uses the maximum mean firing rates among all classes of labeled neurons for inference in the test process.
In our model, we achieved the same algebraic computation and reasoning using an additional layer of spiking neurons instead of the outside-network computations, thus achieving considerable progress in STDP-based SNN models for SL due to the completely spike-based computations. Moreover, there were two other improvements in the USL process in our model compared to that of Diehl and Cook \cite{diehl2015unsupervised}. The first improvement was the novel sym-STDP rule rooted in DA-STDP, with DA-STDP able to give a potential explanation for the SL processes occurring in the brain. That is, we speculated that DA may be involved in the SL process and that local synaptic plasticity could be changed to sym-STDP during the whole training process. With the aid of the forced firing of a supervised neuron by the incoming teacher signal, sym-STDP could establish the relationship between the input and its teacher information after sufficient training. The second improvement was the new dynamic threshold rule in Eq.~\ref{eq:threshold_intrinsic_plasticity}, in which a decay factor for $\alpha$ was introduced, which could significantly improve performance.

It should be noted that several other SNN training models have performed classification tasks in other ways, but their recognition performance was relatively poor \cite{lin2018relative,sporea2013supervised}. Recently, Xu et al. constructed a novel convolutional spiking neural network (CSNN) model using the tempotron as the classifier and attempted to take advantage of both convolutional structure and SNN temporal coding ability \cite{xu2018csnn}. However, their model only achieved a maximum accuracy of 88$\%$ on a subset of the MNIST dataset (Training samples: 500; Test samples: 100) when the network size equaled $1200$; in contrast, our model achieved an accuracy of 91.41$\%$ with $N_\text{hidden}$ equal to $400$ on the full MNIST dataset. This indicates that our model could also work very well under small network size constraints.

Thus, for the above reasons, our proposed sym-STDP based SNN model could solve the lack of bio-plausible and high-performance SNN methods for spike-based SL.
%%%%%%%%%%%%%%%%%%%%%%%%%%%%%%%%%%%%%%%%%%%%%%%%
\begin{table*}[htb]
        \begin{adjustwidth}{0.1in}{0.1in}
               \scalebox{0.64}{
           \begin{tabular}{ llllll }
             \hline
             \textbf{Network model} & \textbf{Learning method} & \textbf{Training type} & \textbf{(Un-)Supervised}  & \textbf{Accuracy} \\
             \hline
             \multicolumn{5}{c}{\textbf{In-direct training }} \\
             \hline
       Deep LIF SNN   & \multirow{2}* {BP-ANN conversion}   & \multirow{2}* {Rate-based}     & \multirow{2}* {Supervised}  & \multirow{2}* {98.37\%} \\
       \cite{hunsberger2015spiking} & & & & \\
       CSNN\cite{diehl2015fast}    & BP-ANN conversion   & Rate-based     & Supervised  & 99.1\% \\
       Chip-based SNN \cite{esser2015backpropagation}   & BP-ANN conversion   & Rate-based   & Supervised   & 99.42\% \\
       SDRN\cite{hu2018spiking}   & BP-ANN conversion   & Rate-based   & Supervised   & 99.59\% \\
             \hline
             \multicolumn{5}{c}{\textbf{Direct training }} \\
             \hline
       BP-STDP SNN\cite{tavanaei2017bp}   & BP-STDP   & Rate-based  &   Supervised   &96.6\% \\
       Deep LIF-BA SNN  & \multirow{2}* {Broadcast Alignment}   &  \multirow{2}* {Rate-based}   &  \multirow{2}* {Supervised}  & \multirow{2}* {97.05\%} \\
       \cite{Samadi2017Deep} & & & &\\
       STBP SNN\cite{wu2017spatio}   & Spatio-Temporal BP   &  Rate-based  & Supervised   &98.89\% \\
       VDPC SNN \cite{zhangaaai2018}    & Equilibrium learning + STDP & Voltage-based & Supervised & 98.52\%  \\
       Deep SNN\cite{lee2016training}   & BP   & Voltage-based  & Supervised   & 98.88\% \\
       \multirow{2}* {DCSNN\cite{Lee2018}}  & \multirow{2}* {STDP + BP} & Spike-based \& & \multirow{2}* {Supervised} & \multirow{2}* {99.28\%} \\
        & &  Current-based & &\\
       Two-layer SNN\cite{querlioz2013immunity}   & Rectangular STDP   & Spike-based   & Unsupervised   & 93.5\% \\
       %Spiking RBM\cite{o2013real}    & Contrastive Divergence   & Probability-based     & Supervised  & 94.1\% \\
       Two-layer SNN \cite{diehl2015unsupervised}   & Exponential STDP    & Spike-based   & Unsupervised   & 95.0\% \\
       SDNN\cite{kheradpisheh2017stdp}   & STDP + SVM  & Spike-based   & Unsupervised   & 98.4\% \\
       Three-layer SNN\cite{shrestha2017stable}  & Specially modified STDP  & Spike-based  & Supervised  & 89.7\% \\
       MLHN\cite{Beyeler2013Categorization}   & STDP with Calcium variable   & Spike-based   & Supervised   & 91.6\% \\
       Bidirectional SNN\cite{hu2017stdp}   & Specially modified STDP     & Spike-based   & Supervised   &96.8\% \\
       DCSNN\cite{mozafari2018combining}  & STDP + R-STDP & Spike-based & Supervised & 97.2\% \\
       sym-STDP SNN (\textbf{Ours})      & sym-STDP (DA-STDP)  & Spike-based     & Supervised & 96.73\%  \\
             \hline
           \end{tabular}
           }
         \caption{Comparison of classification performance between our SNN model and others on the MNIST task. All models reported here were trained based on the full training dataset and tested by the full testing dataset. LIF: Leaky integrate-and-fire; BP: Backpropagation; CSNN: Convolutional spiking neural network; SDRN: Spiking deep residual net- work; MLHN: Multi-layer hierarchical network; STBP: Spatio-temporal backpropagation; VDPC: Voltage-driven plasticity-centric; BA: Broadcast alignment; SDNN: Spiking deep neural network; SVM: Support vector machine; R-STDP: Reward-modulated STDP.}
         \label{tab:SNN-models-comparison}
        \end{adjustwidth}
       \end{table*}
       %% more refs suggested for tabel: MNIST:96.8%---Towards deep learning with segregated dendrites,Elife, 2017(12) 
%%%%%%%%%%%%%%%%%%%%%%%%%%%%%%%%%%%%%%%%%%%%%%%%

\begin{table*}[htb]
        \centering
        \begin{tabular}{c c c}
          \toprule
          Network Size ($N_\text{hidden}$)&400&6400   \\
          \midrule
          Simultaneous SL & $77.61\%\pm0.40\%$ & $84.70\%\pm 0.24\%$ \\
          \midrule
          Layer-by-layer SL & $78.68\%\pm0.27\%$ & $85.31\%\pm 0.16\%$ \\
          \midrule
          Label Stat. & $77.73\%\pm0.50\%$ & $84.89\%\pm 0.21\%$ \\
          \bottomrule
        \end{tabular}
        \caption{Performance on Fashion-MNIST dataset for different sized networks with different training methods. For network size $N_\text{hidden} = 400$ and $6400$, $\tau_{\theta} = 5\times 10^{7}$ and $2\times 10^{7}$, $\alpha = 5\times 10^{6}$ and $2\times 10^{6}$, $\beta = 0.05$ and $0.025$, respectively.}
        \label{tab:perf-for-fashion}
\end{table*}

\section{Robustness of our SL model}
To demonstrate the robustness of our SL model, we also tested its performance on Fashion-MNIST \cite{xiao2017fashion}, a MNIST-like fashion product dataset with 10 classes. Fashion-MNIST shares the same image size and structure of the training and testing splits as MNIST but is considered more realistic as its images are generated from front look thumbnail images of fashion products on Zalando's website via a series of conversions. Therefore, Fashion-MNIST poses a more challenging classification task than MNIST. We preprocessed the data by normalizing the sum of a single sample gray value because of high variance among examples. We then made necessary parameter adjustments to the model mentioned earlier. Specifically, some class examples in Fashion-MNIST have more non-zero pixels than MNIST, such as T-shirt, pullover, shirt, and coat. We decreased $\beta$ in Eq.~\ref{eq:synaptic_scaling} so as to reduce weights and offset the impact of excessive spike quantity. We trained our model under different network sizes ($N_\text{hidden}$ = 400 and 6400). The same evaluation criteria were applied, as shown in Table~\ref{tab:perf-for-fashion}, our model also performed well on the Fashion-MNIST task under both two SL training methods. For example, the Layer-by-Layer training methods achieved accuracies of 78.68$\%$ and 85.31$\%$ for network sizes 400 and 6400, respectively. The best performance of our model is comparable with traditional machine learning methods, such as SVM with linear kernel (83.9$\%$) and multilayer perceptron (the highest accuracy reported was 87.1$\%$) \cite{xiao2017fashion}. These results further confirm the robustness of our SL model. 
\section{Discussion}
A neural network model with biological plausibility must meet three basic characteristics, i.e., the ability to integrate temporal input and generate spike output, spike-based computation for training and inference, and all learning rules rooted in biological experiments. Here, we used the LIF neuron model, with all learning rules (e.g., sym-STDP, synaptic scaling, and dynamic threshold) rooted in experiments and computation based on spikes. Thus, the proposed SNN model meets all the above requirements and is a true biologically plausible neural network model.

However, how did our model obtain good pattern recognition performance? This was mainly because the three learning rules worked synergistically to achieve good feature extraction and generate the appropriate mapping from input to output. The sym-STDP rule demonstrated a considerable advantage by extracting the relationship of spike events, regardless of their temporal order, in two connected neurons, with synaptic scaling able to stabilize total in-synaptic weights and create weight competition among in-synapses of a neuron to ensure that the suitable group of synapses became strong. Furthermore, the dynamic threshold mechanism compelled a neuron to fire for matched patterns but rarely for unmatched ones, which generated neuron selectivity to a special class of patterns. By combining the three bio-plausible plasticity rules, our SNN model established a strong relationship between the input signal and supervised signal after sufficient training, ensuring effective SL implementation and good performance in the benchmark pattern recognition task (MNIST). The proposed model also obtained good performance by training two layers synchronously, whereas many previous SNN models require layer-by-layer or multi-phase/multi-step training \cite{kheradpisheh2017stdp,lee2016training,Samadi2017Deep,zhangaaai2018,hu2017stdp}.
% we proposed a novel bio-plausible SNN model for SL based on the sym-STDP rule. The sym-STDP is rooted in the observed DA-STDP rule in experiments, where dopamine makes the classic STDP rule lost its temporal order information and become a sym-STDP rule\cite{Zhang2009Gain,Ruan2014,brzosko2015retroactive}. 

To explore the effect of different deep network structures, we also tested the performance of our SL algorithm when adding or removing a hidden layer in our network model. Taking $N_{\textup{E}} = 100$ as an example, the network model with no hidden layer achieved a performance of only 57.74$\%$, whereas the network model with two hidden layers reached 81.65$\%$, slightly lower than that achieved in the network with one-hidden layer (83.57$\%$). In deep ANNs, classification performance is usually improved by increasing the number of hidden layers; however, this phenomenon was not found in our model. In our model, each hidden layer excitatory neuron has a global receptive field, which differs from that in deep convolutional neural networks (CNNs), where each convolutional layer neuron has a common local receptive field. A CNN needs more convolutional layers to generate larger receptive fields and thus achieve a global receptive field in the last layer; however, this is not required in our model. After training, the global receptive field of a hidden-layer neuron resulted in a maximum response of the neuron to a special category of input patterns, thus allowing hidden-layer neurons to be easily divided into different classes. That is, unsupervised clustering can be realized using just one hidden layer. A subsequent supervised layer can then directly implement classification after training based on supervised label information. Therefore, in the framework of our network model, it would be useless to promote performance by increasing hidden layers, and also would harm performance by removing the hidden layer as the model can only generate 10 special global receptive fields (by 10 supervised neurons) for all 60000 training patterns. Therefore, in our model, we increased the number of hidden-layer neurons to improve performance rather than the number of hidden layers. As no more than 10 neurons usually responded to an input pattern in the hidden layer, activity was so sparse that simple linear mapping from the high-dimension activities in the hidden layer to the low-dimension activities in the classification layer was possible, again indicating no need for further hidden layers. Nevertheless, additional hidden layers may be necessary to improve performance if the convolutional structure (local receptive fields) is adopted for the network. We will explore convolutional SNNs to further verify the universality of our SL algorithm in the future.

In our SNN model, DA was found to be a key factor for achieving SL. Dopamine plays a critical role in different learning processes and can serve as a reward signal for reinforcement learning \cite{glimcher2011understanding,holroyd2002neural,wise2004dopamine,dayan2002reward}. A special form of DA-modulated STDP, different to the symmetric one used here, has been applied for reinforcement learning in SNNs previously \cite{izhikevich2007solving}; however, no direct experimental evidence for this kind of STDP rule has yet been reported. Here, we assumed that DA may also be involved in the SL process, with the symmetric DA-STDP rule found in experiments to modify synaptic weights during SL. Our work further indicated the potentially diverse functions of DA in regulating neural networks for information processing. Moreover, even if this DA-involved assumption is shown to be biologically unsound in the future, direct experimental evidence, i.e., identification of a sym-STDP without the need of DA in the hippocampal CA3-CA3 synapses under slow-frequency stimulation~\citep{mishra2016symmetric}, supports the bio-plausibility of the sym-STDP rule used in our model. It is worth noting that the sym-STDP rule is a spiking version of the original Hebbian rule, that is, `cells that fire together wire together', and a rate-based neural network with the Hebbian rule, synaptic scaling, and dynamic bias could be expected to have similar classification abilities as our model. However, the performance of the rate model may not be as high as that reported here. Further exploration is needed using the simplified rate model with the original Hebbian rule.

Several SNN models have been developed to provide a general framework for SNN learning, e.g.,~liquid state machine (LSM) \citep{maass2002real} and NeuCube \citep{kasabov2014neucube,kasabov2018time}. The LSM is a classic model of reservoir computing (RC)~\citep{jaeger2001echo,jaeger2004harnessing,lukovsevivcius2009reservoir,lukovsevivcius2012reservoir,maass2002real}, which contains an input module, a reservoir module consisting of relatively large spiking neurons with randomly or fully connected structure, and a readout module. In a typical LSM, the low dimensional input, usually possessing spatiotemporal information, is fed into the high-dimensional space of the reservoir to optimally maintain a high-dimensional, dynamic representation of information as a short-term memory when the state of the reservoir is at the edge of chaos. The reservoir is then projected to the readout module where synaptic weights are modified by learning methods. NeuCube is also a general SNN model similar to a RC model, with both a similar network structure and ability to encode spatiotemporal input as LSM~\citep{kasabov2014neucube,kasabov2018time}. In NeuCube, the reservoir is a three-dimensional-based structure and can dynamically evolve its weight structure using unsupervised learning rules, which allows it to encode an input as a special activity trajectory in the high-dimension activity space of the reservoir. In the readout module of NeuCube, an evolving SNN (eSNN) is usually adopted for classification~\citep{kasabov2018time}, which can dynamically create a new output neuron for clustering by evaluating the similarity between a new input and encoded inputs in Euler space. Our model is a general SNN model for SL. There are some significant differences between our model and the two models. Firstly, our learning rule (sym-STDP) is consistent for different layers across the whole model, whereas NeuCube generally uses different learning rules for synapses in the reservoir and out-synapses reservoir and the classic LSM only uses a learning rule to adjust the weights of readout synapses from the reservoir with fixed inner connections. Secondly, our model contains biologically plausible receptive fields, which can be learned during training, whereas the concept of receptive fields is ambiguous in NeuCube and LSM. Compared to the two models, however, our model lacks the ability to process complex temporal information. Therefore, in the future, it would be useful to construct a model possessing the advantages of the above models to improve the ability to process complex spatiotemporal information.

As the plasticity rules used here were based purely on local spike events, in contrast with the BP method, our model not only has the potential to be applied to other machine learning tasks under the SL framework but may also be suitable for online learning on programmable neuromorphic chips. Moreover, our hypothesis regarding the function of DA in SL processing may serve as a potential mechanism for synaptic information processing of SL in the brain, which will need to be verified in future experiments.

The code is available at https://github.com/haoyz/sym-STDP-SNN.

\section*{Acknowledgments}
\begin{sloppypar}
This work was supported by the National Natural Science Foundation of China (Grant No. 11505283), Beijing Brain Science Project (Grant No. Z181100001518006), and Strategic Priority Research Program of the Chinese Academy of Sciences (Grant No. XDB32070000). We also gratefully acknowledge the support of the NVIDIA Corporation for the donation of the Titan X Pascal GPU used in this research.
\end{sloppypar}
%\section*{References}

\bibliography{DASTDP_SL_REF_R2_abbriev}

\end{document}